\documentclass{article}
\usepackage{spconf,amsmath,graphicx}
\usepackage{amsmath,graphicx}
\usepackage{amsfonts}
\usepackage{color}
\usepackage{caption}
\usepackage{subfigure}
\usepackage{amsthm}

\newtheorem{thm}{Theorem}

\newtheorem{prop}[thm]{Proposition}

\include{definitions}
\title{Recovery of noisy points on bandlimited surfaces: kernel methods re-explained}
\name{Sunrita Poddar, Mathews Jacob\thanks{This  work  is  supported  by NIH 1R01EB019961-01A1.}}
\address{Department of Electrical and Computer Engineering, University of Iowa, IA, USA}
\begin{document}
%
\maketitle
\begin{abstract}
We introduce a continuous domain framework for the recovery of points on a surface in high dimensional space, represented as the zero-level set of a bandlimited function. We show that the exponential maps of the points on the surface satisfy annihilation relations, implying that they lie in a finite dimensional subspace. The subspace properties are used to derive sampling conditions, which will guarantee the perfect recovery of the surface from finite number of points. We rely on nuclear norm minimization to exploit the low-rank structure of the maps to recover the points from noisy measurements.  Since the direct estimation of the surface is computationally prohibitive in very high dimensions, we propose an iterative reweighted algorithm using the "kernel trick". The iterative algorithm reveals deep links to Laplacian based algorithms widely used in graph signal processing; the theory and the sampling conditions can serve as a basis for discrete-continuous domain processing of signals on a graph.
\end{abstract}

\section{Introduction}
The recovery of signals that lie on a manifold/surface has received extensive attention in the recent years. For example, patch-based image processing methods such as BM3D model the patches in an image as points on a manifold \cite{bm3d,yasir}, while we \cite{storm, poddar2014joint} have recently used the manifold structure of images in a dynamic time series. Another area that witnessed extensive research in the recent years is the processing of signals on a graph \cite{graph}; these methods rely on the the graph Laplacian operator to denoise and post-process signals on a graph. 

The main focus of this paper is to introduce a continuous domain perspective on the recovery of points drawn from a smooth surface in very high dimensions. This work reveals fundamental links between recent advances in superresolution theory \cite{candes,gregpapers,recht} and kernel based machine learning methods \cite{scholkopf} as well as graph signal processing \cite{spm}. We assume that the high dimensional points live on an smooth surface, which is the zero level set of a bandlimited function. This is termed the annihilation relation and it is shown that this relation can be expressed as a weighted linear combination of the exponential features of the point; the dimension of the feature maps is equal to the bandwidth of the potential function. These properties enable us to determine the sampling conditions, which will guarantee the recovery of the surface from finite number of points. Our analysis also shows that when the bandwidth is overestimated, there are multiple such annihilation relations, suggesting that the exponential feature maps of the points on the surface live in a finite dimensional space. Note that similar non-linear maps are widely used in kernel methods; our results show that these maps can be approximated by a few basis functions, when the points are restricted to a bandlimited surface.

The finite dimensional nature of the maps translate to a low-rank kernel matrix, computed from the points using a shift invariant kernel such as the Dirichlet function. We minimize the nuclear norm of the feature maps of the points to recover them from noisy data. Since the direct estimation  of the surface in higher dimensions suffers from the curse of dimensionality, we use the "kernel trick" to keep the computational complexity manageable. We rely on an iterative reweighted algorithm to recover the denoised points. The resulting algorithm has similarities to iterative non-local methods \cite{wendy,yasir,mohsin2015iterative,osher,graph} that are widely used in image processing and graph signal processing. Specifically, it alternates between the estimation of a graph Laplacian, which specifies the connectivity of the points, and the smoothing of points guided by the graph Laplacian. Our experiments show that the Laplacian matrix obtained by solving the proposed optimization algorithm is more representative of the graph structure than classical methods, when it is estimated from noisy data.

This work is built upon our prior work \cite{ongie2015recovery, gregpapers, ongie2017convex, balachandrasekaran2017recovery, balachandrasekaran2017novel, ongie2017fast} and the recent work by Ongie et al., which considered polynomial kernels \cite{gregvariety}. Our main focus is to generalize \cite{gregvariety} to shift invariant kernels, which are more widely used. We also introduce sampling conditions and algorithms to determine the surface, when the dimension is low. In addition, the iterative algorithm using the kernel trick shows the connections with graph Laplacian based methods used in graph signal processing. 

\section{Bandlimited surfaces \& annihilation}
\begin{figure}[t!]
\centering
\includegraphics[width=0.48\textwidth]{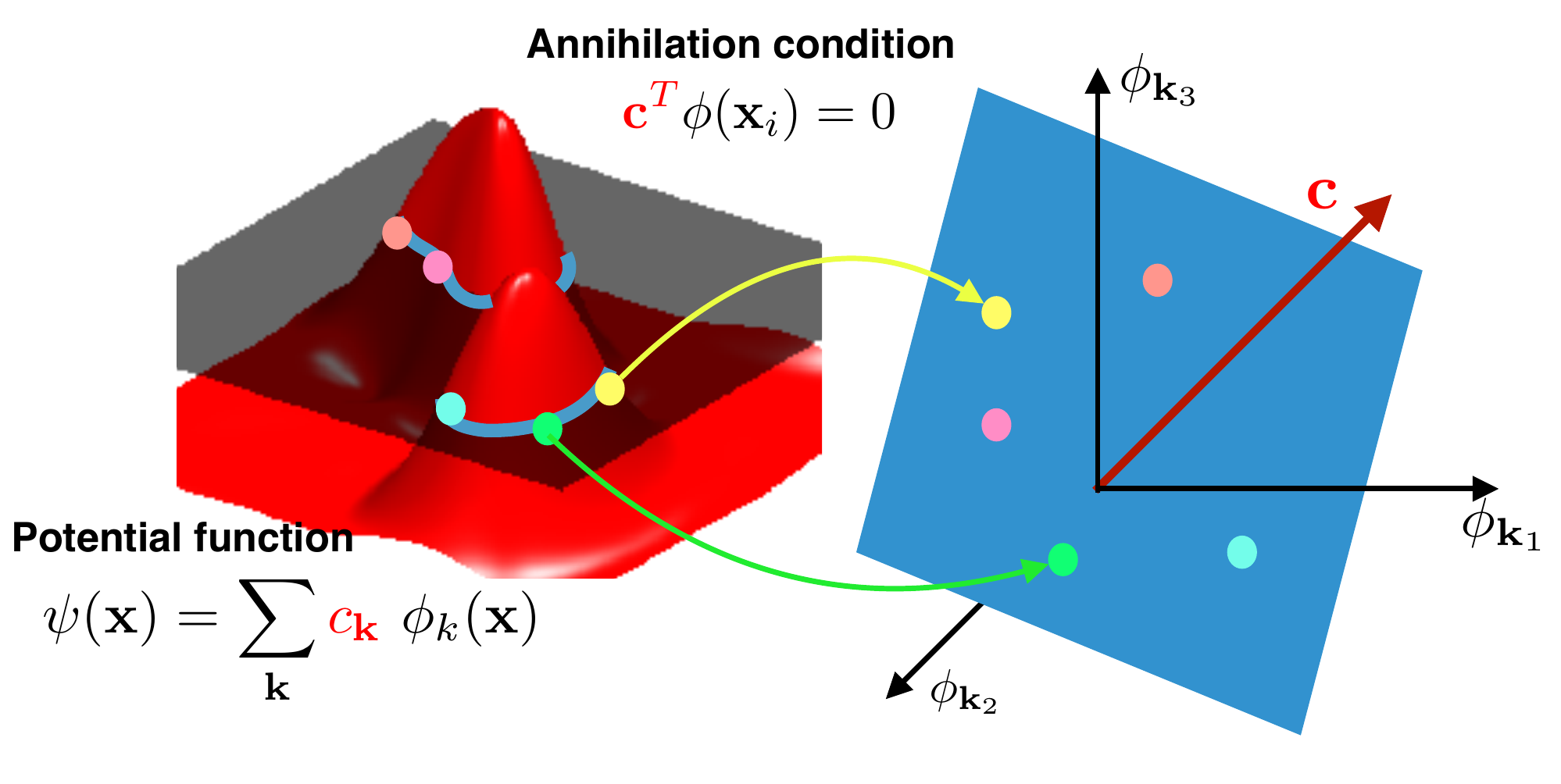}

\caption{Illustration of the annihilation relations in 2-D. We assume that the curve is the zero-level set of a bandlimited function $\psi(\mathbf x)$. Each point on the curve satisfies $\psi(\mathbf x_i)=0=\mathbf c^T\phi_{\Lambda}(\mathbf x_i)$, which can be seen as an annihilation relation in the non-linear feature space $\phi_{\Lambda}(\mathbf x)$. Specifically, the maps of the points lie on a plane orthogonal to $\mathbf c$.}
\label{illus}
\end{figure}

We assume the point cloud to be supported on a surface in $[-1/2,1/2]^n$, which is the zero level-set of a bandlimited potential function:
\begin{equation}
\label{implicit}
\{\mathbf r \in \mathbb R^n|\psi(\mathbf r)=0\}, ~\mbox{where}~\psi(\mathbf r) = \sum_{\mathbf k \in \Lambda} \mathbf c_k \exp(j~2\pi\mathbf k^T \mathbf r)
\end{equation}
Here, $\{\mathbf c_{\mathbf k}: \mathbf k\in \Lambda\}$ is the smallest set of coefficients (minimal set) that satisfies the above relation. $\Lambda\subset \mathbb Z^{n}$ is a set of contiguous locations that indicates the support of the Fourier series coefficients of $\psi$. Consider an arbitrary point $\mathbf x$ on the above surface \eqref{implicit}. By definition \eqref{implicit}, we have the annihilation relation $\psi(\mathbf x) = \sum_{\mathbf k \in \Lambda} \mathbf c_k \exp(j~2\pi\mathbf k^T \mathbf x) = 0$. We re-express the annihilation relation as $\mathbf c^T\phi_{\Lambda}(\mathbf x) = 0$ using a non-linear mapping $\phi_{\Lambda}: \mathbb R^n \rightarrow \mathbb C^{|\Lambda|}$:
\begin{equation}
\phi_{\Lambda}(\mathbf x) = \begin{bmatrix} \exp(j~2\pi\mathbf k_1^T\mathbf x)&
 \ldots&
  \exp(j~2\pi\mathbf k_{|\Lambda|}^T\mathbf x)
\end{bmatrix}^T
\end{equation}
This annihilation relation is illustrated in Fig \ref{illus}. 

\subsection{Curve recovery: sampling conditions}
The annihilation relation introduced in the previous sub-section can be used to estimate the surface, or equivalently $\psi(\mathbf r)$ from a few number of points. The least square estimation of the coefficients from the data points $\{\mathbf x_i: i=1,\ldots,N\}$ can be posed as the minimization of the criterion: 
\begin{equation}
\mathcal C(\mathbf c) = \sum_{i=1}^N \|\psi(\mathbf x_i)\|^2= \mathbf c^T \mathbf Q_{\Lambda} \mathbf c
\end{equation}
where $\mathbf Q_{\Lambda}=\sum_{i=1}^N \phi_{\Lambda}(\mathbf x_i)\phi_{\Lambda}(\mathbf x_i)^T$. The coefficients can be estimated as:
\begin{equation}
\label{eigen}
\mathbf c^* = \arg \min_{\mathbf c} \mathbf c^T ~\mathbf Q_{\Lambda}~ \mathbf c ~~\mbox{such that }~~ \|\mathbf c\|^2 = 1
\end{equation}
The solution is the minimum eigen vector of $\mathbf Q_{\Lambda}$. 

In the remainder of the section, we will restrict our attention to 2-D for simplicity, even though the results in this section can be generalized to arbitrary dimensions. We will now determine the sampling conditions for the perfect recovery of the curve $\psi(\mathbf x)=0$ using \eqref{eigen}. Specifically, we will determine the minimum number of samples for the successful recovery of the curve, when $\Lambda$ is a rectangular neighborhood in $\mathbb Z^2$ of size $K_1\times K_2$. In addition, we assume that $\psi$ is the function with the smallest Fourier support (minimal polynomial), whose zeros define the curve. We first focus on the case where $\Lambda$ is known. 
\begin{prop}
Let $\mathbf x_i; i=1,..,N$ be points on the zero-level set of a band-limited function $\psi(\mathbf r), \mathbf r \in \mathcal R^2$, where the bandwidth of the surface $\psi$ is specified by $|\Lambda| = K_1\times K_2$ and $\psi(\mathbf r)$ has $J$ irreducible factors. If $N_j$ points are sampled on the $j^{th}$ irreducible factor, then the curve $\psi(\mathbf r) = 0$ can be uniquely recovered by \eqref{eigen}, when:
\begin{equation}
\label{nsamples}
N_j> (K_1+K_2)(K_1^{j}+K_2^{j})
\end{equation}
for $j=1,\ldots,J$.
\end{prop}
Thus, the total number of points required are $N > (K_1+K_2)(K_1+K_2+2(J-1))$. We compare this setting with the sampling conditions for the recovery of a piecewise constant image, whose gradients vanish on a bandlimited curve \cite{gregpapers}. The minimum number of Fourier measurements required to recover the function there is $|3\Lambda|$; when $K_1=K_2=K$, then $3K^2$ complex Fourier samples are required. In contrast, we need $4K^2$ real samples. When the true support $\Lambda$ is not known, it is a common practice to overestimate it as $\Gamma \supset \Lambda$. In this case, $\mathbf Q_{\Gamma}$ will have multiple null space vectors, as shown below.
\begin{prop}
We consider the polynomial $\psi(\mathbf r)$ described in Proposition 1. Let $\Lambda \subset \Gamma$ with $|\Gamma| = L_1\times L_2$ and for $j=1,\ldots,J$:
\begin{equation}
\label{overSamp}
N_j> (L_1+L_2)(K_1^{j}+K_2^{j})
\end{equation}
 points be sampled on the $j^{th}$ irreducible factor of $\psi(\mathbf r)$. Then all nullspace vectors $\mathbf c' \stackrel{\mathcal F}{\leftrightarrow} \psi'$ of the matrix $\mathbf Q_{\Gamma}$ will be of the form: 
\begin{equation}
\psi'(\mathbf r) = \psi(\mathbf r)\;\eta(\mathbf r)
\end{equation}
where $\eta(\mathbf x)$ is an arbitrary function such that $supp(\mathbf c') = \Gamma$. 
\end{prop}
Thus, the total number of points required are $N > (L_1+L_2)(K_1+K_2+2(J-1))$.
Since $\psi(\mathbf x)$ is the common factor of all the annihilating functions, all of them will satisfy $\psi'(\mathbf x)=0$, for any point on the original curve. Depending on the specific $\eta$, they will have additional zeros. Hence, the above result provides us a means to compute the original curve, even when the original bandwidth/support of the function is unknown. 

\begin{figure}[t!]
\centering
\center{\includegraphics[width=0.5\textwidth]{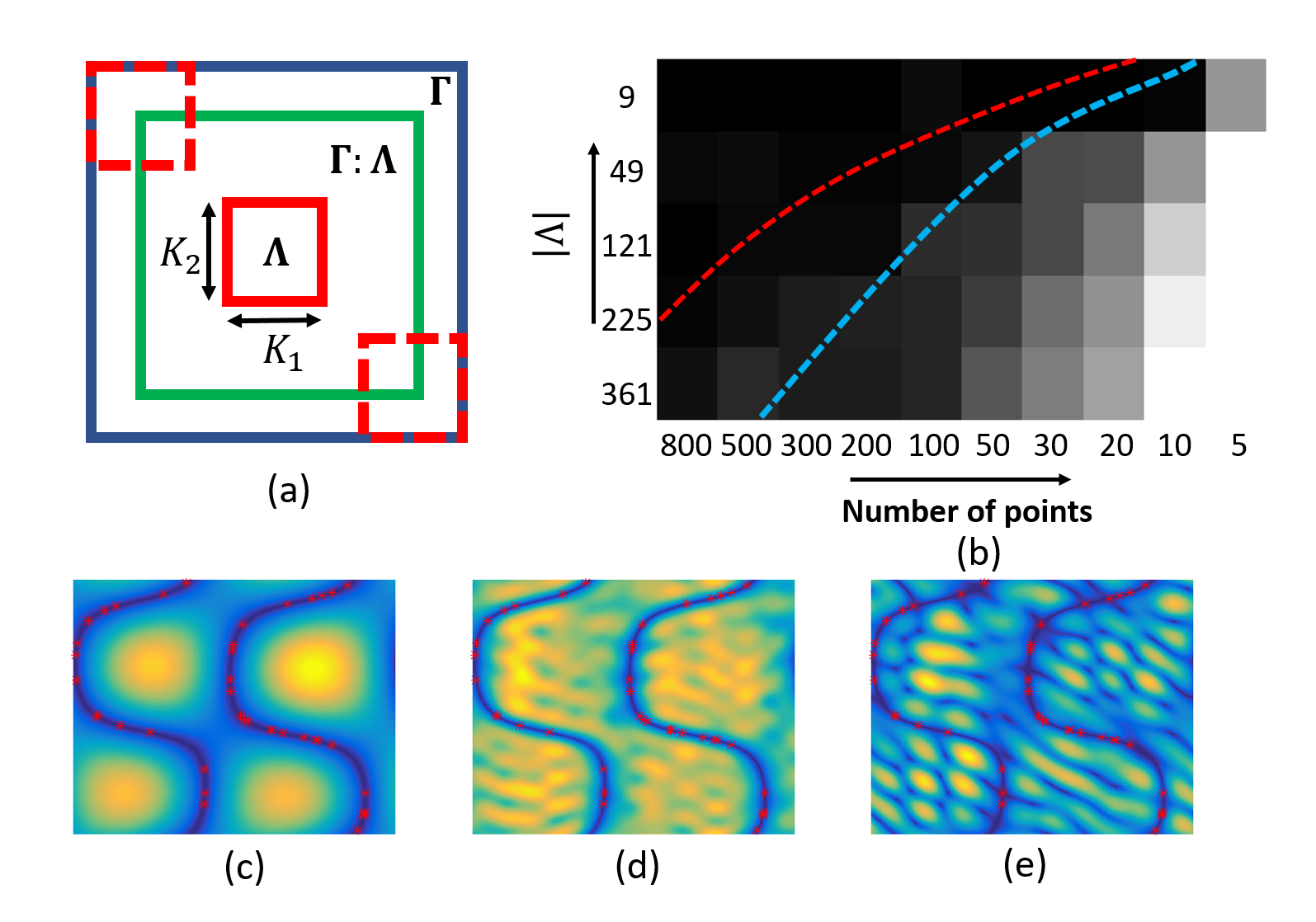}}
\caption{Sampling conditions: The Fourier support $\Lambda$ of the minimal function $\psi$, the overestimated support $\Gamma$ used to evaluate the maps, and the possible shifts of $\Lambda$ in $\Gamma$ denoted by $\Gamma:\Lambda$ are shown in (a). (b) shows the phase transition plots, where the red curve is the one predicted by the theory, and the blue curve is $N = |\Lambda|$. Here, black indicates perfect recovery and white denotes poor recovery. The recovery of the curve with $|\Lambda|=9$ in (c) from its points denoted by red points is illustrated in (d)-(e). We assumed $\Gamma$ to be a $11\times 11$ region. (e) shows one null space filter. The sum of square of $10$ null space filters in (d) uniquely identifies the curve.}
\label{phasetrans}
\end{figure}

We now consider a collection of $N $ points on the curve, stacked into a matrix $
\mathbf X = \left[\mathbf x_1,\mathbf x_2,\ldots \mathbf x_N\right]$. Let the feature matrix of size $|\Gamma|\times N$ be denoted by:
\begin{equation}
\label{matrixannihilation}
\Phi_{\Gamma}(\mathbf X) = \begin{bmatrix}
\phi_{\Gamma}(\mathbf x_1),\ldots \phi_{\Gamma}(\mathbf x_N)
\end{bmatrix}
\end{equation}
We state a result about the rank of the above feature matrix.
\begin{prop}
We consider the polynomial $\psi(\mathbf r)$ described in Proposition 1 and $\Lambda \subset \Gamma$. Then:
\begin{equation}
\label{rankbound}
{\rm rank}\left(\Phi_{\Gamma}(\mathbf X) \right) \leq \underbrace{ |\Gamma|-|\Gamma:\Lambda|}_r
\end{equation}
with equality if the sampling conditions of Proposition 2 are satisfied.
\end{prop}
Here, $|\Gamma:\Lambda|$ denotes the number of valid shifts of the set $\Lambda$ within $\Gamma$ as shown in Fig \ref{phasetrans} (a). Note that as $|\Lambda|$ gets smaller, the number of shifts of it within $\Gamma$ increases, and hence the rank decreases. The rank of the matrix can be used as a surrogate for the bandwidth of $\psi$, or equivalently the complexity of the curve. Note that $\Lambda$ may be an irregular shape in $\mathbb Z^n$. For example, if the points lie on a line in $\mathbb R^n$, then $\Lambda$ could be concentrated along a line in $\mathbb Z^n$, resulting in a small $|\Lambda|$, even when the number of features in $\Gamma$ may be considerably high. The low-rank structure of the feature maps can be used to denoise the original points, while the sum of squares function obtained from the nullspace filters can be used to estimate the surface in low-dimensions when \eqref{overSamp} is satisfied, as illustrated in Fig \ref{phasetrans}. 

\subsection{Recovery of noisy point clouds in high dimensions}
The explicit approach of estimating the surface is feasible, when the dimension of the points $n$ is small. However, this approach suffers from the curse of dimensionality. Since the shape of the data, or equivalently the shape of the support $\Lambda$ is not known, one needs to use a large $\Gamma$ to ensure that $\Lambda \subset \Gamma$. Note that the dimension of the feature space specified by $|\Gamma|$ grows exponentially with $n$, making this approach impractical in applications involving point clouds of images or patches. 

We hence rely on the right nullspace relations to recover the points from their noisy and undersampled measurements. Specifically, we are interested in the null space relations
\begin{equation}
\underbrace{\Phi_{\Gamma}(\mathbf X)^H\Phi_{\Gamma}(\mathbf X)}_{\mathbf K^{\Gamma}}\mathbf v_i = \mathbf 0
\end{equation}
where the entries of the $|N|\times |N|$ Gram matrix $\mathbf K_{\Gamma}$ are 
\begin{equation}
\label{kernel}
\left(\mathbf K_{\Gamma}\right)_{i,j} = \phi_{\Gamma}(\mathbf x_i)^H\phi_{\Gamma}(\mathbf x_j) =\underbrace{ \sum_{\mathbf k\in \Gamma}  \exp\left(j~2 \pi\mathbf k^T \left(\mathbf x_j-\mathbf x_i\right)\right)}_{\kappa_{\Gamma}(\mathbf x_j-\mathbf x_i)}
\end{equation}
The function $\kappa_{\Gamma}(\mathbf r)$ in \eqref{kernel} is shift invariant and is dependent on the shape of $\Gamma$. For example, when $\Gamma$ is a centered cube in $\mathbf R^n$, $\kappa_{\Gamma}(\mathbf r) $ is a Dirichlet function. The kernel matrix satisfies  ${\rm rank}(\mathbf K_{\Gamma}) \leq r$, where $r$ is given by \eqref{rankbound}. 
  
\subsection{Dirichlet and Gaussian surface representation}
The bandlimited function $\psi(\mathbf r)$ in \eqref{implicit} can equivalently be expressed as: 
\begin{equation}
\psi(\mathbf r) = \sum_{\mathbf l\in \Gamma^c } d_{\mathbf l} ~\varphi_{\Gamma}(\mathbf r-\mathbf l)
\end{equation}
where $\varphi_{\Gamma}(\mathbf x)$ is the Dirichlet function dependent on $\Gamma$ and $\Gamma^c$ is the set of sampled locations on the curve. Using reciprocity, the non-linear maps in this case can be shown to be: 
\begin{equation}
\phi_{\Gamma}(\mathbf x) = \begin{bmatrix} \varphi_{\Gamma}(\mathbf x-\mathbf x_1)&
    \ldots&
   \varphi_{\Gamma}(\mathbf x-\mathbf x_{|\Gamma^c|})
\end{bmatrix}^T
\end{equation}
Since the implicit curve is the zero level set of a linear combination of Dirichlet functions, it may be highly oscillatory.  An alternative would be to use a level set expansion in terms of weighted exponentials $\exp(-\pi^2 \sigma^2\frac{\|\mathbf k\|^2}{2}).\exp(j2\pi \mathbf k^T\mathbf r)$, which could give smoother surfaces. In this case $\kappa_{\Gamma}$ approaches a periodized Gaussian function, as $\Gamma\rightarrow \mathbb Z^n$, and the Gaussian kernel matrix $\mathbf K_{\Gamma}$ is theoretically full rank. However, we observe that the Fourier series coefficients of a Gaussian function can be safely approximated to be zero outside $|\mathbf k|< 3/\pi\sigma$, which translates to $|\Lambda| \approx \left(\frac{6}{\pi\sigma}\right)^n$; i.e., the rank will be small for high values of $\sigma$. We choose Gaussian kernels since they are more isotropic and less oscillatory than the Dirichlet kernel.

\begin{figure}[t!]
\centering
\center{\includegraphics[width=0.4\textwidth]{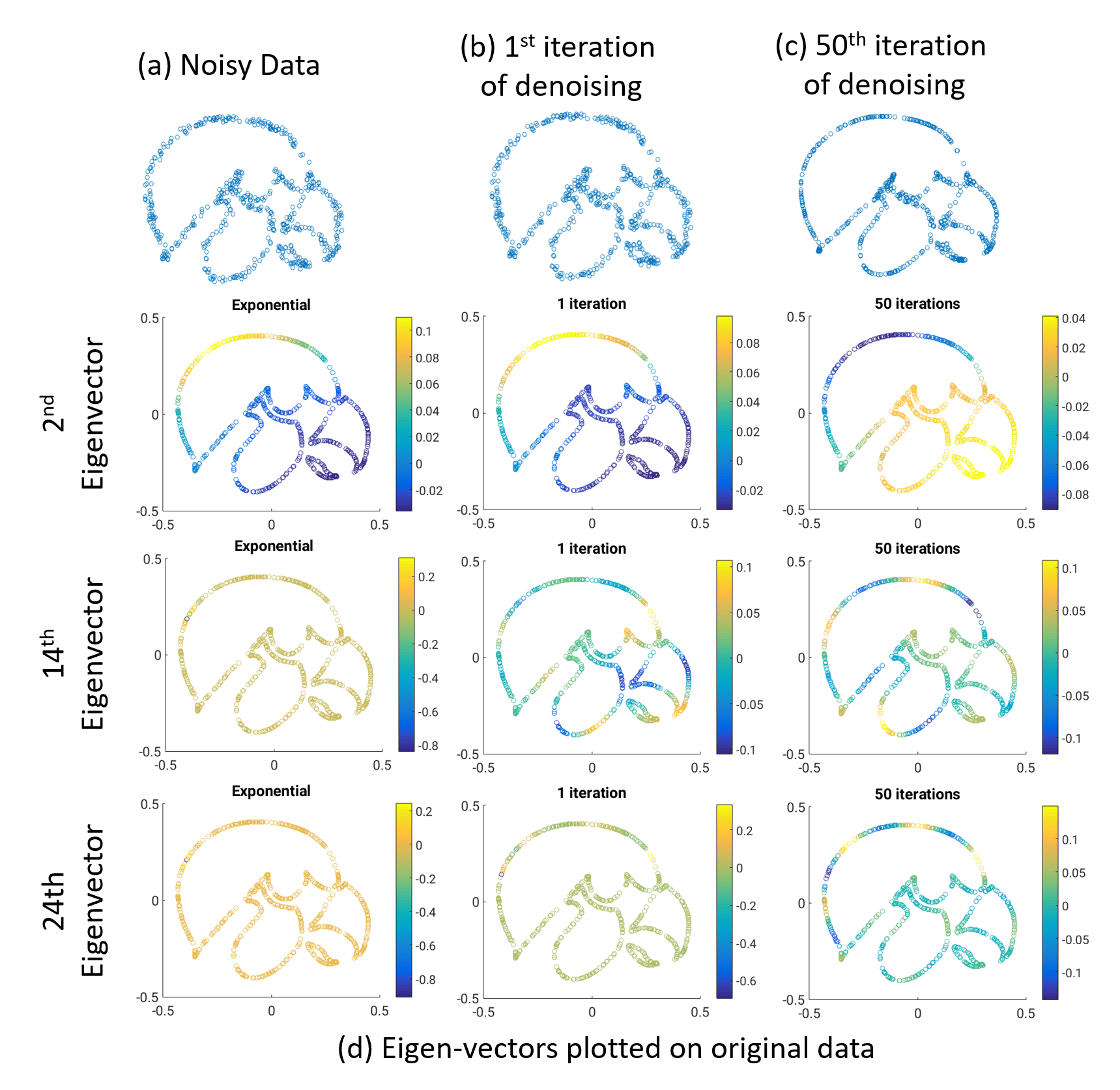}}
\caption{Illustration of denoising of 2-D points on a curve using \eqref{opt}: The top row denotes the noisy data, the first iteration of \eqref{sub1}, and the $50^{th}$ iterate respectively. Note that the kernel low-rank algorithm provides good recovery of the points with 50 iterations. The algorithm also provides a robust approach to estimating the Laplacian from noisy data. The three columns correspond to the eigen vectors of the Laplacians (analogous to Fourier exponentials) estimated from the noisy data using Gaussian kernels, first iteration of the algorithm, and the $50^{th}$ iterate, respectively. We observe that all Laplacian estimation schemes provide good estimates of the 2nd eigen vector, while only the iterative strategy is able to provide good estimates of the higher ones (e.g. bottom row), demonstrating the benefit of the proposed scheme. }
\label{Figherky}
\end{figure}

\subsection{Denoising using nuclear norm minimization}
We rely on the low rank structure of the kernel matrix $\mathbf K$ to recover the noisy points. Specifically, with the addition of noise, the points deviate from the zero set of $\psi$. A high bandwidth potential function is needed to represent the noisy surface. We propose to use the nuclear norm of the feature matrix as a regularizer in the recovery of the points from noisy measurements:
\begin{equation}
\label{opt}
\mathbf X^* = \arg\min_{\mathbf X} \|\mathbf X - \mathbf Y\|^2 + \lambda\|\mathbf \Phi(\mathbf X)\|_*
\end{equation}
We use the IRLS algorithm, where $\mathbf X$ is updated as:
\begin{equation}
\label{sub1}
\mathbf X^{(n)} = \arg \min_{\mathbf X} \|\mathbf X - \mathbf Y\|^2 + \lambda~ {\rm trace}[\mathbf K(\mathbf X)\mathbf Q^{(n)}]
\end{equation}
and $\mathbf Q^{(n)} = [\mathbf K(\mathbf X^{(n-1)}) + \gamma^{(n)} \mathbf I]^{-\frac{1}{2}}$. Note that the solution for \eqref{sub1} involves a system of non-linear equations. Instead, we use gradient linearization to simplify our computations, where $\mathbf K(\mathbf X)$ is a Gaussian kernel matrix: 
\begin{equation}
\label{sub1new}
\mathbf X^{(n)} = \arg \min_{\mathbf X} \|\mathbf X - \mathbf Y\|^2 + \lambda ~{\rm trace}(\mathbf X^T \mathbf L^{(n)} \mathbf X)
\end{equation}
with $\mathbf L^{(n)} = \mathbf D^{(n)} - \mathbf W^{(n)}$, $\mathbf D^{(n)}_{ii} = \sum_j \mathbf W^{(n)}_{ij}$, and 
\begin{equation}
\label{wts}
\mathbf W^{(n)} = -\frac{1}{\sigma^2}{\mathbf K}(\mathbf X^{(n-1)})\odot\mathbf Q^{(n)}
\end{equation}

We note the equivalence of the above optimization strategy with widely used non-local means and graph optimization schemes. These schemes estimate a Laplacian matrix $\mathbf L$, followed by the minimization of the cost function \eqref{sub1new}. These approaches can thus be seen as fitting a smooth bandlimited surface to the point cloud of patches or signals that are assumed to be on the graph. 

\section{Results}
We demonstrate the utility of \eqref{opt} in a simple 2-D denoising example in Fig \ref{Figherky}. Specifically, we consider the recovery of points on  the TigerHawk logo from its noisy samples. See the caption for details. The top row shows that the proposed algorithm is able to provide good denoising of the data. The bottom three rows show that the Laplacian estimated using \eqref{wts} at the $50^{th}$ iteration is more representative of the shape. 

\begin{figure}[t!]
	\centering
	\center{\includegraphics[width=0.48\textwidth]{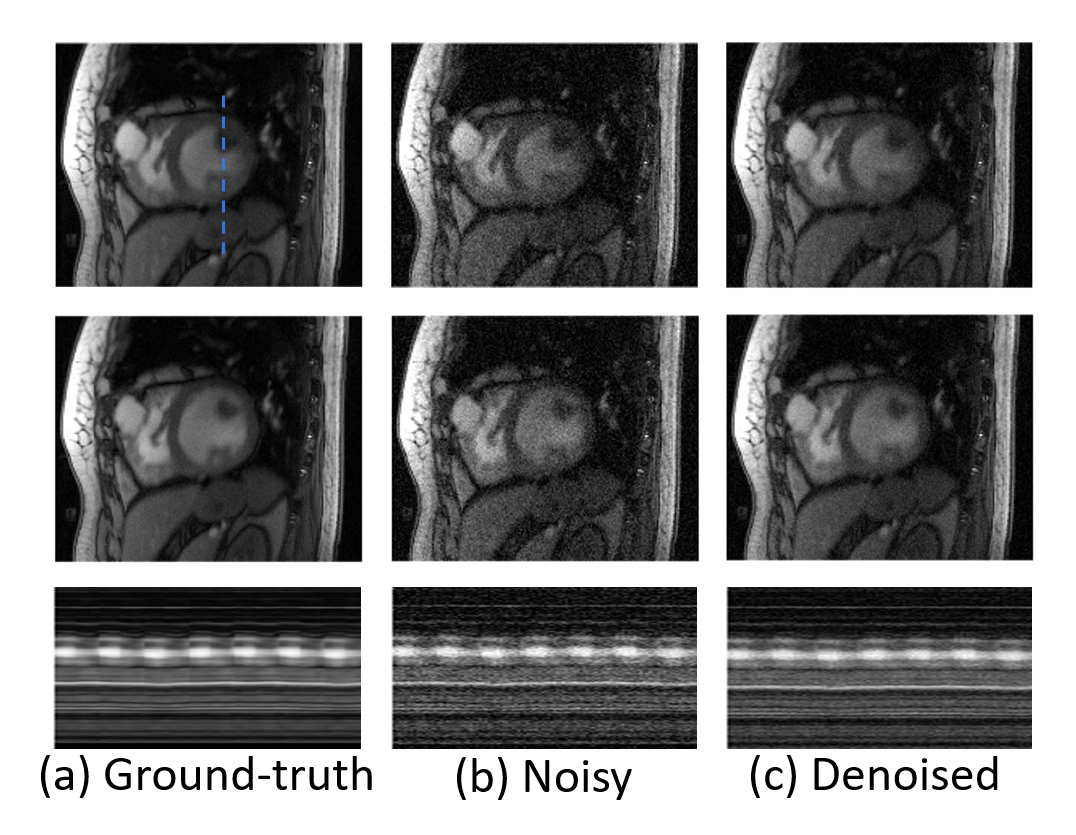}}
	\caption{Denoising a free breathing and ungated cardiac MRI image series: (a), (b) and (c) show the ground-truth, noisy data and denoised data respectively. Out of the 200 frames, two selected image frames are shown along with the temporal profile along the blue line. }
	\label{FigCard}
\end{figure}

The utility of the proposed method in denoising free breathing and ungated MRI data is shown in Fig \ref{FigCard}. Since MRI is a slow imaging modality, several rapid imaging techniques were introduced to accelerate the acquisition. All of these methods trade SNR for speed, resulting in noisy images. The proposed scheme is seen to exploit the manifold structure of the data to reduce noise. See caption for details.

\section{Conclusion}
We introduce a continuous domain framework for the recovery of points on a bandlimited surface. We show that the exponential maps of the points lie in a lower dimensional subspace, which translates to a kernel matrix that is low-rank. We introduce a nuclear norm minimization algorithm to recover the points. The proposed framework connects the continuous domain surface recovery problem with kernel methods and approaches in graph signal processing. The application of the algorithms to noisy data reveals its great utility in practical problems. 

\bibliographystyle{IEEEbib}
\bibliography{strings,refs}

\end{document}